\pdfoutput=1
\documentclass[runningheads,a4paper]{llncs}

\usepackage{microtype}
\usepackage{amssymb}
\usepackage{graphicx}
\usepackage{url}
\usepackage{wrapfig}

\usepackage[colorlinks=true,linkcolor=black,citecolor=black,urlcolor=black]{hyperref}

\graphicspath{{images/}}
\DeclareGraphicsExtensions{.pdf,.jpg,.png}

\usepackage{eso-pic}

\AtBeginDocument{\AddToShipoutPictureFG*{\AtTextUpperLeft{\put(0,\LenToUnit{40pt}){\parbox{\textwidth}{\centering\bfseries
RoboCup 2013: Robot World Cup XVII, Lecture Notes in\\Computer Science 8371, pp.\ 56-67, Springer, 2014
}}}}}
\begin{document}

\mainmatter  %

\title{Learning to Improve Capture Steps\\ for Disturbance Rejection in Humanoid Soccer}
\titlerunning{Learning to Improve Capture Steps for Winning Humanoid Soccer}

\author{Marcell~Missura, Cedrick~M\"unstermann, Philipp~Allgeuer, Max~Schwarz,
Julio~Pastrana, Sebastian~Schueller, Michael~Schreiber, and Sven~Behnke}
\authorrunning{M. Missura, C. M\"unstermann, P. Allgeuer, et al.}
\institute{%
Autonomous Intelligent Systems, Computer Science, Univ.\ of Bonn, Germany\\
\path|{missura, schreiber}@ais.uni-bonn.de|,~
\path|behnke@cs.uni-bonn.de|\\
\url{http://ais.uni-bonn.de}}

\maketitle

\begin{abstract}
Over the past few years, soccer-playing humanoid robots have advanced significantly.
Elementary skills, such as bipedal walking, visual perception, and collision
avoidance have matured enough to allow for dynamic and exciting games. 
When two robots are fighting for the ball, they frequently push each other and
balance recovery becomes crucial. In this paper, we report on insights we
gained from systematic push experiments performed on a bipedal model and
outline an online learning method we used to improve its push-recovery capabilities. 
In addition, we describe how the localization ambiguity introduced by the
uniform goal color was resolved and report on the results of the RoboCup 2013
competition.

\end{abstract}

\begin{figure}[b!]
\centering
	\includegraphics[height=3.7cm]{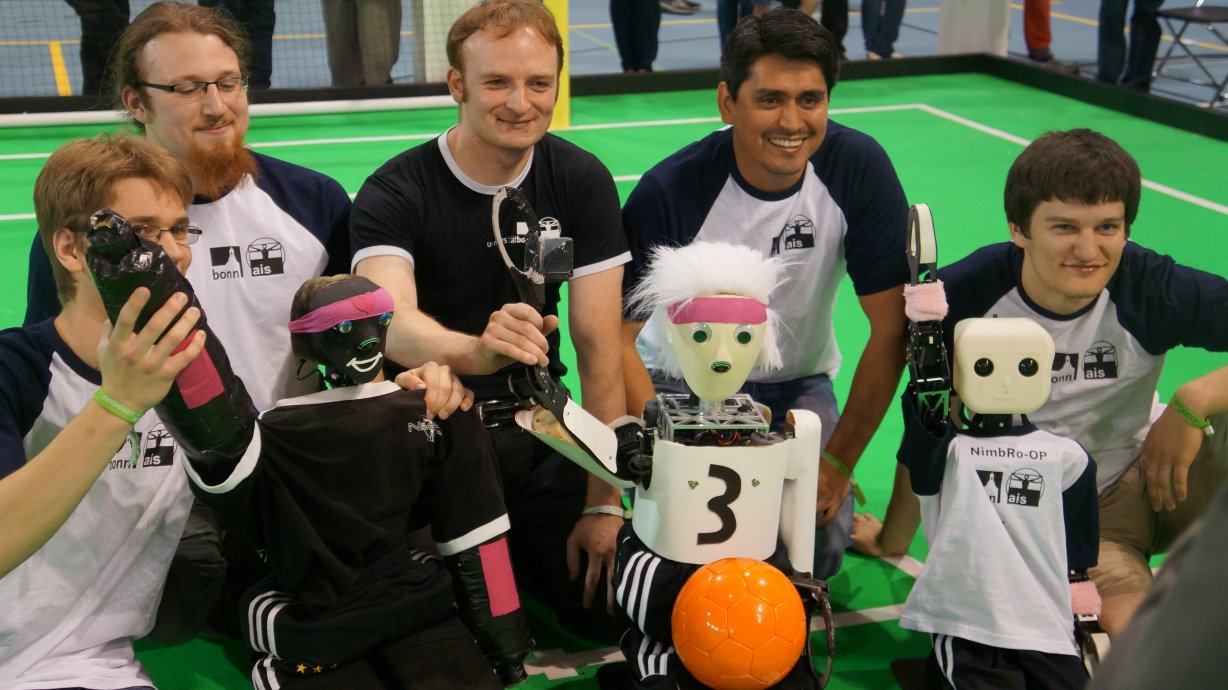}\hspace*{2ex}
	\includegraphics[height=3.7cm]{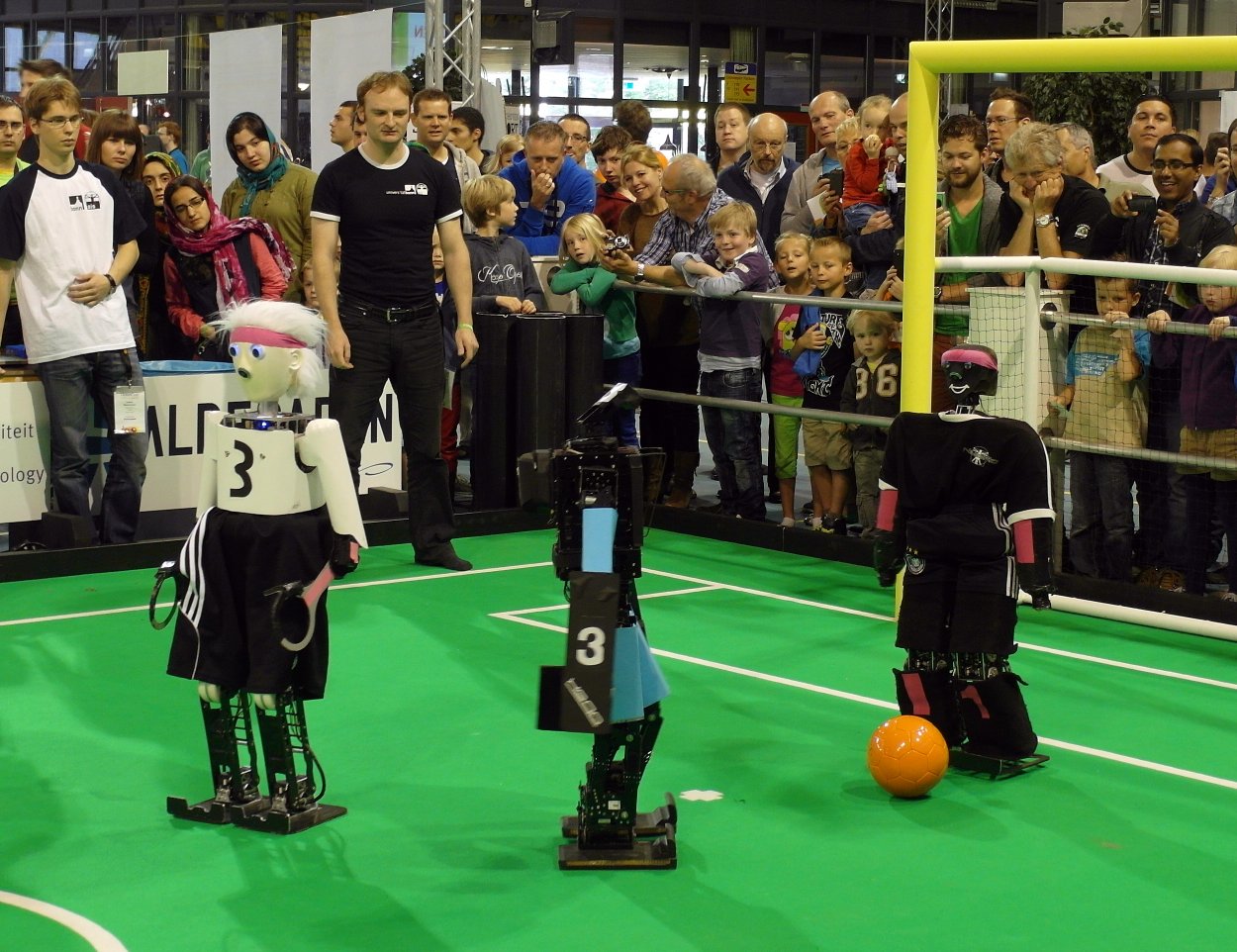}\vspace*{-1ex}
	\caption{Left: Team NimbRo with robots Dynaped, Copedo, and NimbRo-OP. Right:
	Team NimbRo vs. CIT-Brains in the RoboCup 2013 finals.}
	\label{nimbro:bots}
\end{figure}
\vspace{-1em}

\section{Introduction}

In the RoboCup Humanoid League, robots with a human-like body plan compete against
each other in soccer games. The robots are largely self-constructed, and are
divided into three size classes:
KidSize ($<$60\,cm), TeenSize (90--120\,cm), and AdultSize ($>$130\,cm). The
TeenSize robots started to play 2 vs. 2 soccer games in 2010 and moved to a
larger soccer field of 9$\times$6\,m in the year 2011. In addition to the
soccer games, the robots face technical challenges, such as throwing the ball
into the field from a side line.

For RoboCup 2013, the color coding of the goal posts was unified to yellow for
both goals and the landmark poles at the ends of the center line were removed.
Consequently, it was not possible anymore to determine the unambiguous position
of a robot on the field based only on visual cues, which constitutes a problem
for localization. However, most teams were able to implement suitable solutions
and were able to reliably drive the ball towards the opponent goal. Our approach
to disambiguate localization was to integrate a compass as an additional source
of information. More details are given in Section~\ref{chap:perception}.

Inspired by the success of the \mbox{DARwIn-OP} robot, we have constructed a
TeenSize open platform, the NimbRo-OP. Following the same spirit, the NimbRo-OP
is a low-cost robot that is easy to construct, maintain, and extend. It is
intended to provide access to a humanoid robot platform for research. The
NimbRo-OP has matured enough to participate in the competitions. It participated
in the Technical Challenges and scored its first official competition goal in the main event.
More information about the NimbRo-OP is given in Section~\ref{chap:nop}.

Bipedal walking is a crucial skill in robot soccer. It determines the success of
a team to a substantial degree. Humanoid robots must be able to walk up to a
ball and kick it, preferably without losing balance and falling to the ground.
While most of the teams have mastered the skill of unperturbed walking on flat
terrain, solutions to recover from strong disturbances, such as collisions with
opponents, are not yet widespread. In ongoing research, team NimbRo has
developed a stable bipedal gait control framework that has been designed to
absorb strong perturbations. In Section~\ref{chap:learning}, we report on the
insights we gained from systematic push experiments, and introduce an online
learning method that we used to improve push recovery capabilities. The learning
controller is able to adjust the step size and recover balance quicker than the
underlying simplified mathematical model.

\section{Mechatronic Design of NimbRo TeenSize Robots}

The mechatronic design of our robots is focused on robustness, weight reduction,
and simplicity. All our robots are constructed from milled carbon fiber and
aluminum parts that are assembled to rectangular shaped legs and flat arms. We
use Dynamixel EX-106 and EX-106+ servos for the actuation of our classic robots
Dynaped and Copedo. These robots are also equipped with spring-loaded protective
joints that yield to mechanical stress and can snap back into place
automatically. More information about the mechanical structure of the NimbRo
classic robots can be found in \cite{winners2012} and \cite{winners2011}. The
NimbRo-OP robot has a slightly different design with a reduced complexity. It is
equipped with 6\,DOF legs and 3\,DOF arms that offer enough flexibility to walk,
to kick, and to get up from the floor after falling. It is actuated by servos
from the Dynamixel MX series. The mechatronic structure of the NimbRo-OP is best
described in \cite{Schwarz2012}.

\section{Perception}
\label{chap:perception}

\begin{figure}[b]
\centering
\includegraphics[height=2.4cm]{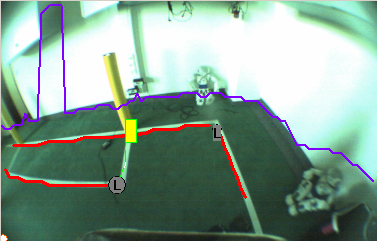}\hspace*{3ex}
\includegraphics[height=2.4cm]{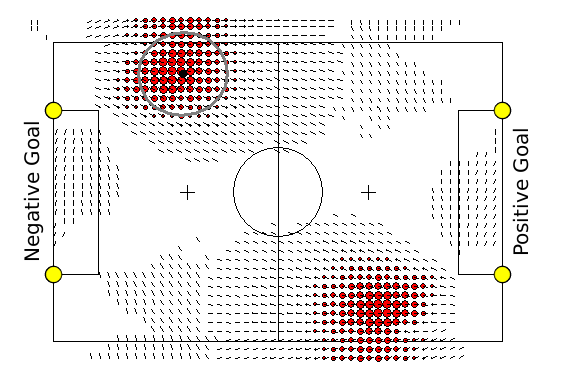}\hspace*{3ex}
\includegraphics[height=2.4cm]{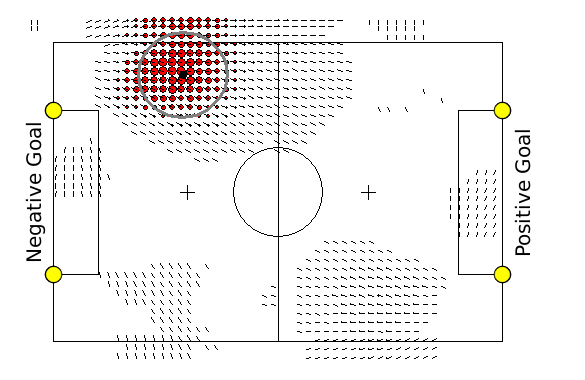}\vspace*{-1ex}
\caption{Effect of the compass on localization confidence. The observed scene
in the camera image (left) leads to two hypothesis peaks in the particle
distribution of the particle filter (center). Adding the compass reading as an
additional observation disambiguates the position estimation (right).}
\label{compasspf}
\end{figure}

For visual perception of the game situation, we detect the ball, goal-posts,
penalty markers, field lines, corners, T-junctions, X-crossings, obstacles, team
mates, and opponents utilizing color, size and shape information. We estimate
distance and angle to each detected object by removing radial lens distortion
and by inverting the projective mapping from field to image plane.

For proprioception, we use the joint angle feedback of the servos and apply it
to the kinematic robot model using forward kinematics. Before extracting the
location and the velocity of the center of mass, we rotate the kinematic model
around the current support foot such that the attitude of the trunk matches the
angle we measured with the IMU. Temperatures and voltages are also monitored for
notification of overheating or low batteries.

For localization, we track a three-dimensional robot pose $(x, y, \theta)$ on
the field using a particle filter~\cite{probabilisticrobotics}. The particles
are updated using a linear motion model. Its parameters are learned from motion
capture data~\cite{Schmitz:Footstep}. The weights of the particles are updated
according to a probabilistic model of landmark observations (distance and angle)
that accounts for measurement noise. To handle unknown data association of
ambiguous landmarks, we sample the data association on a per-particle basis. The
association of field line corner and T-junction observations is simplified using
the orientation of these landmarks. Further details can be found in
\cite{Schulz:lines} and \cite{winners2011}.

\noindent {\bf Integration of a compass:} This year, we extended our sensory
systems with a compass in order to help the particle filter to disambiguate the
localization on the field. As starting from 2013 both goals have the same
color and there are no landmarks that allow unambiguous localization based only
on visual cues, it was necessary to add an additional source of information
other than the objects detected by the computer vision. Using the compass output
as observation of the global orientation in the particle filter greatly helps
to reduce the number of hypothesis that can accumulate in the particle
distribution. Figure~\ref{compasspf} shows such an example. The robot observes a
situation in the corner of the field, where field lines, L-shaped line crossings
and a goal post have been successfully detected. Despite the high number of
observations that the particles can be weighted with, two equally valid
hypotheses form, as shown by the particle distribution in the center. Adding the
global heading as additional observation reduces the probability of particles
that are facing in a wrong direction. Thereby one of the hypothesis in this
example is invalidated (right). As an additional benefit of using a compass, we
found that it not only improves localization, but also the effectiveness of our
soccer behaviors. This is due to the fact that the rough direction of the
opponent goal is always known. Thus, the ball is always moved in the right
direction, even in cases where the particle filter reports a wrong pose.

\section{Behavior Control}

We control our robots using a layered framework that supports a hierarchy of
reactive behaviors~\cite{Behnke:Hierarchical}. When moving up the hierarchy, the
update frequency of sensors, behaviors, and actuators decreases, while the level
of abstraction increases. Currently, our implementation consists of three
layers. The lowest, fastest layer is responsible for generating motions, such as
walking \cite{MissuraBehnkeWalking} ---including capture
steps~\cite{Missura:LateralCaptureSteps}, kicking, get-up motions
\cite{Stueckler:GetUp}, and the goalie dive \cite{Missura:Goalie}. At the next
higher layer, we model the robot as a simple holonomic point mass that is
controlled with the force field method to generate ball approach trajectories,
ball dribbling sequences, and to implement obstacle avoidance. The topmost layer
of our framework takes care of team behavior, game tactics and the
implementation of the game states as commanded by the referee box. Please refer
to \cite{winners2011} for further details.

\section{NimbRo-OP TeenSize Robot}
\label{chap:nop}

\begin{wrapfigure}{r}{0.32\textwidth}
\centering
\vspace*{-5ex}
\includegraphics[width=0.28\textwidth]{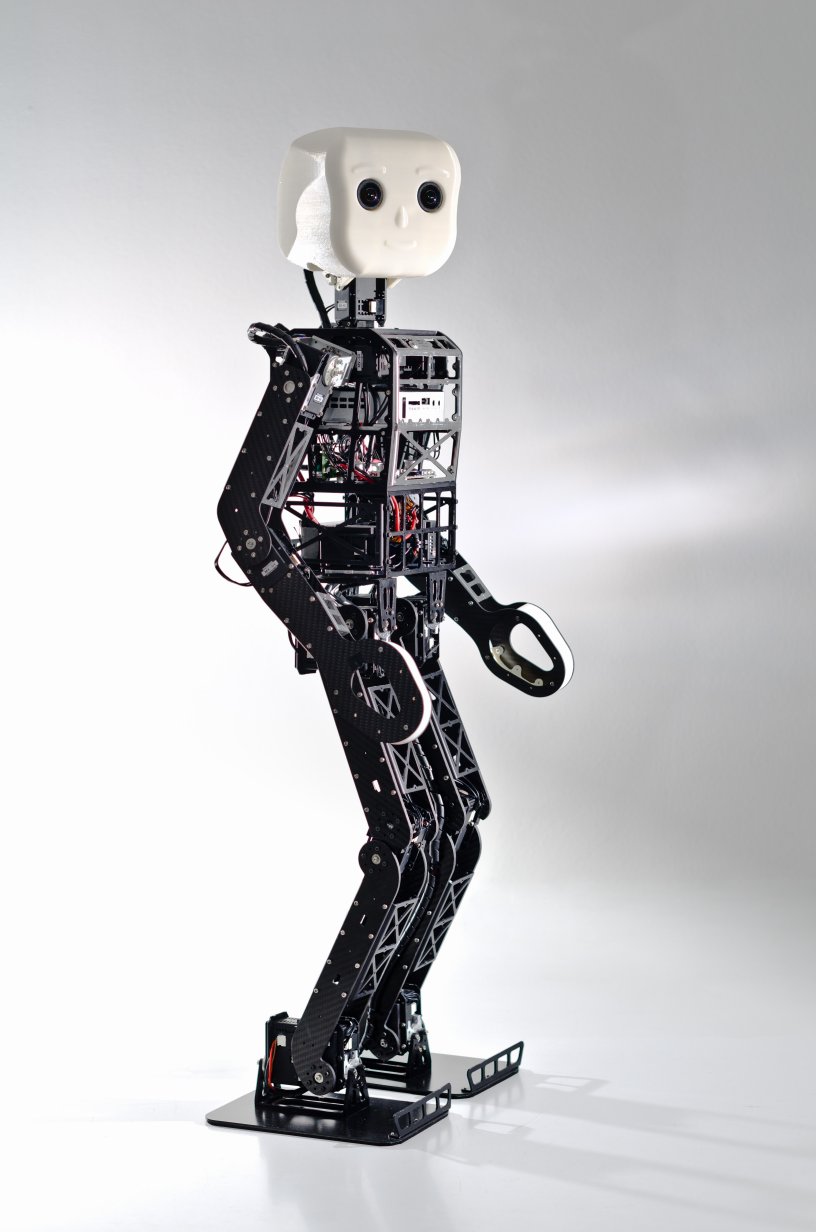}\vspace*{-1ex}
\caption{The \mbox{NimbRo-OP}.}
\label{nop}
\vspace*{-5ex}
\end{wrapfigure}

Our main innovation this year was the development of the NimbRo-OP robot along
with a ROS framework based robot soccer software. The software contains many
modules for basic functions required for playing soccer that we either started
from scratch, or ported from our classic NimbRo system. In the now second
release \cite{Allgeuer2013a}, the software package contains a compliant servo
actuation module \cite{MaxCompliantControl} and a visual motion editing
component. Motions are replayed with a non-linear keyframe interpolation
technique that allows to generate smooth and continuous motions while respecting
configurable acceleration and velocity bounds. Kicking and get-up motions have
been successfully implemented. For walking, we use a port of the same gait
generator that we use for our classic robots \cite{MissuraBehnkeWalking}. For
higher-level behavior control, we ported the NimbRo hierarchical reactive
behavior architecture \cite{Behnke:Hierarchical} \cite{Allgeuer2013} and the
implementations of simple soccer behaviors within, such as searching for the
ball, walking up to the ball and dribbling the ball. The vision processing
module was rewritten from scratch as a ROS module along with accompanying tools
for camera and color calibration. Utilizing a camera with higher resolution and
more available processing power, we improved the quality of our object
detection, which is described in \cite{Allgeuer2013a} in more detail. A particle
filter-based localization module is also provided. Apart from the core soccer
software itself, graphical software components are available to maintain
configuration parameters and to log the state of the system in great detail to
support debugging and monitoring during games.

\section{Online Learning of Lateral Balance}
\label{chap:learning}

In recent years, team NimbRo has developed a gait control framework capable of
recovering from pushes that are strong enough to force a bipedal walker to
adjust step-timing and foot-placement. Only lateral balance mechanisms
\cite{Missura:LateralCaptureSteps} have been used in competitions so far, but in
simulation, the framework is now able to absorb pushes from any direction at any
time during the gait cycle \cite{Missura:OmnidirectionalCaptureSteps}. In a
nutshell, the Capture Step Framework is based on an extremely simplified state
representation in the form of a point mass that is assumed to behave like a
linear inverted pendulum. A decomposition of the lateral and sagittal dimensions
into independent entities, and a sequential computation of step-timing,
zero-moment point and foot-placement control parameters facilitates the
closed-form mathematical expression of our balance controller. Modeling,
however, can only take one so far. Complex full-body dynamics, sensor noise,
latency, imprecise actuation, and simplifying modeling assumptions will always
result in errors that can limit the balancing capabilities of a humanoid robot.
A good way to increase the efficiency of a model based approach are
online learning techniques that can measure performance during walking and
adjust the output of model-based push-recovery strategies.

Focusing on the simplified purely lateral setting, we have successfully
implemented an online learning algorithm that learns the foot-placement error
during disturbed walking on the spot and subtracts it from the model output in
order to improve push recovery capabilities. In the following section, we briefly
outline the concepts of lateral balance and introduce our evaluation method that
can quantify and visualize the effects of isolated balance components. Subsequently,
we describe the online learning algorithm we used, and show experimental
results to verify the achieved improvement.

\subsection{Lateral Gait Control}

\begin{figure}[h] \centering
\includegraphics[width=0.95\textwidth]{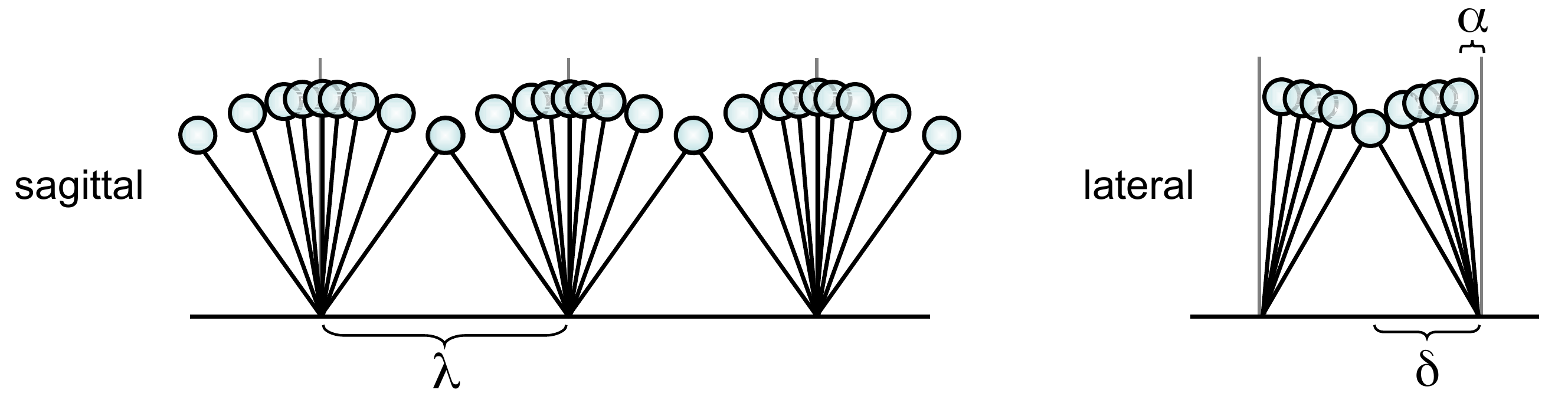} 
\vspace*{-2ex}
\caption{Stick diagrams of idealized pendulum-like sagittal and lateral motion
of a compass gait. In sagittal direction, the center of mass crosses the pendulum
pivot point in every gait cycle, while in lateral direction it oscillates
between the support feet. Parameter $\lambda$ defines the stride length in the
sagittal direction, parameter $\alpha$ denotes the characteristic lateral apex
distance, and $\delta$ defines the support exchange location in the center of
the step.}
\label{Stickfigure}
\end{figure}

The pendulum-like dynamics of human walking has been long known to be a
principle of energy-efficient locomotion~\cite{Kuo:Energy}.
Figure~\ref{Stickfigure} shows stick diagrams of the idealized sagittal and
lateral pendulum motions projected on the sagittal plane and the frontal plane.
Interestingly, the sagittal and lateral motions exhibit strongly distinct
behaviors. In the sagittal plane, the center of mass vaults over the pivot
point in every gait cycle, while in the frontal plane, the center of mass
oscillates between the support feet and never crosses the pendulum pivot point.
It is crucial not to tip over sideways, as the recovery from such an unstable
state requires challenging motions that humanoid robots have difficulties
performing. 

The perpetual lateral oscillation of the center of mass appears to be the
primary determinant of step timing. Disobeying the right timing can quickly
destabilize the system after a disturbance, even if the disturbance itself would
not have directly resulted in a fall \cite{CaptureStepVideo}. Furthermore, we
can identify two characteristic parameters in the lateral direction. We denote the
minimal distance between the pivot point and the center of mass and that
occurs at the apex of the step as $\alpha$. The apex distance provides a certain margin
for error. While during undisturbed walking the apex distance stays near
$\alpha$ in every step, a push in the lateral direction can result in a smaller
apex distance. As long as the apex distance is greater than zero, the
center of mass will return and the walker will not tip over the support foot.
Sooner or later, returning center of mass trajectories are guaranteed to reach
the support exchange location that we denote as $\delta$. While the support
exchange location varies with increasing lateral walking velocity, for now we
limit our setting to walking on the spot with zero velocity of locomotion and
therefore we can assume $\delta$ to be a constant as well. To identify the model
parameters $\alpha$ and $\delta$ for a real or a simulated biped, we induce the
lateral oscillation by generating periodic, open-loop step motions using the
walk algorithm described in \cite{MissuraBehnkeWalking}. Then, $\alpha$ and
$\delta$ can be found by averaging the measured center of mass locations at the
step apex and in the moment of the support exchange.

As a consequence of the principles described above, we can formulate the following
control laws for our balance control computations:

\begin{itemize}
\item The timing of the step is determined by the moment when the
center of mass reaches the nominal support exchange location $\delta$.
\item The lateral step size is chosen so that the center of mass will pass the
following step apex with a nominal distance $\alpha$ with respect to the pivot point. 
\end{itemize}

Formally, our balance controller is a function
\begin{equation}
(T,F) = \mathcal{B}(y, \dot{y})
\end{equation}
that computes the step time $T$ and the footstep location $F$ as a function of the
current state of the center of mass $(y, \dot{y})$. Here, $y$ denotes the
location of the center of mass along the lateral axis with respect to a
right hand coordinate frame placed on the support foot, and
$\dot{y}$ is the velocity of the center of mass. The step time $T$
and the footstep location $F$ are passed on to a motion generator that
generates stepping motions with an appropriate frequency and leg swing
amplitude. For the understanding of the experiments performed in this work, a
conceptual insight of the lateral control laws presented above is sufficient.
For more detailed information, we refer the reader to
\cite{Missura:OmnidirectionalCaptureSteps}.

\subsection{Experimental Setup}

Using a physical simulation software, we performed a series of systematic push
experiments on a simulated humanoid robot with a total body weight of 13.5\,kg
and a roughly human-like mass distribution. While the robot is walking on the
spot, it is pushed in the lateral direction with an impulse targeted at the
center of mass. After the impulse, the robot has some time to recover,
before the next impulse is generated. If the robot falls, it is reset to a
standing position and it is commanded to start walking again. The magnitude of
the impulse is randomly sampled from the range $[-9.0, 9.0]$\,Ns, where the
sign of the impulse determines its direction (left or right). We generate 400
pushes for each of four balance controllers of increasing complexity:
\begin{itemize}
  \item {\bf No Feedback}: The controller ignores the pushes and does
  nothing. The robot executes an open-loop gait with a fixed frequency and step
  size.
  \item {\bf Timing}: The controller adjusts only the timing of the step,
  but not the footstep location.
  \item {\bf Timing + Step Size}: The controller adjusts the timing and the
  size of the steps using the mathematical model.
  \item {\bf Timing + Step Size + Learning}: The controller responds to
  the disturbances using not only the model-based computation of the
  timing and the step size, but also a learned error that we subtract from the
  predicted step size. The error is learned online during the experiment.
\end{itemize}

The input space we use for learning is the lateral state space $\mathcal{S} =
[y, \dot{y}] \in \mathbb{R}^2$ of the center of mass. When the support foot is
the left foot, we flip the signs of $y$ and $\dot{y}$ in order to exploit
symmetry. During the experiment, the robot measures the efficiency of its steps
and estimates an error that expresses a gradient, i.e. a desired scalar
increase or decrease in the step size. The error is measured when the center of
mass is at the step apex. It is given as simply the deviation from the nominal apex
distance $\alpha$. From the inverted pendulum model it follows intuitively that
if the apex distance is greater than $\alpha$, the step size was too large, and
if the apex distance is smaller than $\alpha$, the step size was too small. At
the end of the step, we update the value of a function approximator for each of
the states $(y, \dot{y})_{i \in I}$ that were encountered during the step. The
update rule is
\begin{equation}
f((y, \dot{y})_i) = f((y, \dot{y})_i) + \eta (y_{i_a} - \alpha), i \in I,
\end{equation}
where $f((y, \dot{y})_i)$ is the value of the function approximator for the
state $(y, \dot{y})_i$, $y_{i_a}$ is the center of mass location that was
measured at the step apex, and $\eta = 0.2$ is the learning rate. The function
approximator is initialized with a value of 0 before learning. The step
parameters that are passed on to the step motion generator are then
\begin{equation}
(T,F) = \mathcal{B}(y, \dot{y}) - (0, f(y, \dot{y})).
\end{equation}

\subsection{Evaluation of Results}

\begin{figure}[t]
\centering
\includegraphics[width=0.9\textwidth]{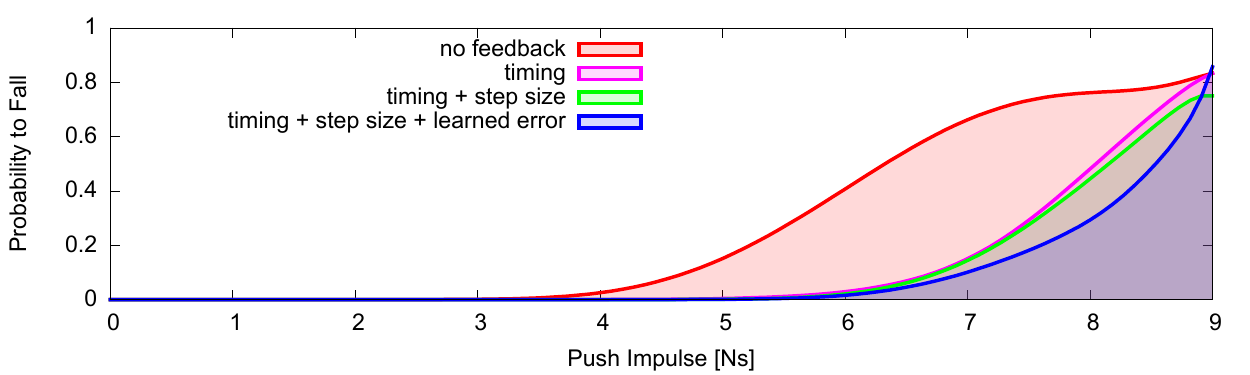}\vspace*{-2ex}
\caption{Probability to fall versus the magnitude of the push impulse for four
different controllers of increasing complexity.}
\label{fallprobability}
\end{figure}

Using the data we collected during the experiments, we can compare the
efficiency of the four controllers. Figure~\ref{fallprobability} shows the
probability to fall against the magnitude of the impulse and gives an impression
of the push resistance of the controllers. Interestingly, the open-loop walk
alone is able to handle pushes up to a strength of 3\,Ns, in such a case
returning slowly to a limit cycle. However, the three feedback controllers
clearly increase the minimum impact required to make the robot fall and improve
the ability to absorb an impact over the entire range of impulse strengths. The
results of the three feedback controllers do not differ from each other
significantly, leading to the conclusion that using the right step timing is
already sufficient to predominantly stabilize returning center of mass
trajectories. Why this effect can be achieved with step timing alone
has a reasonable explanation. When the robot receives a push from the side, it
typically first tilts towards the support leg and the center of mass approaches
the outer edge of the support foot. If the robot was pushed in the direction
away from the support leg, it will automatically tip onto the other leg in the
center of the step, which leads to the same situation. Now, when the center of
mass is moving towards the outer edge of the support foot, the robot may shorten
the support leg if it does not adjust the motion timing, as internally the
support leg is thought to be the swing leg at that time. This accelerates the
center of mass additionally towards the support leg and reduces the lever arm,
helping the robot to tip over the outer edge of the foot. Furthermore, the robot
is likely to touch the floor with the other foot and can further accelerate
itself in the wrong direction. And finally, if the center of mass returns, and
it is moving away from the support leg, a badly timed extension of the support
leg just before the support exchange adds energy to the lateral motion and
increases the probability to tip over on the other side. Using adaptive timing,
all of these undesired effects vanish. The adaptation of step timing prevents
the robot from destabilizing itself due to badly timed leg motions in oblique
poses and maximizes the minimal tip-over impulse to the value that can be
passively absorbed. Using the torso as a reaction mass for active balancing
could further increase the minimal tip-over impulse, but this is not in our
scope at this time.

\begin{figure}[t]
\centering
\includegraphics[width=0.49\textwidth]{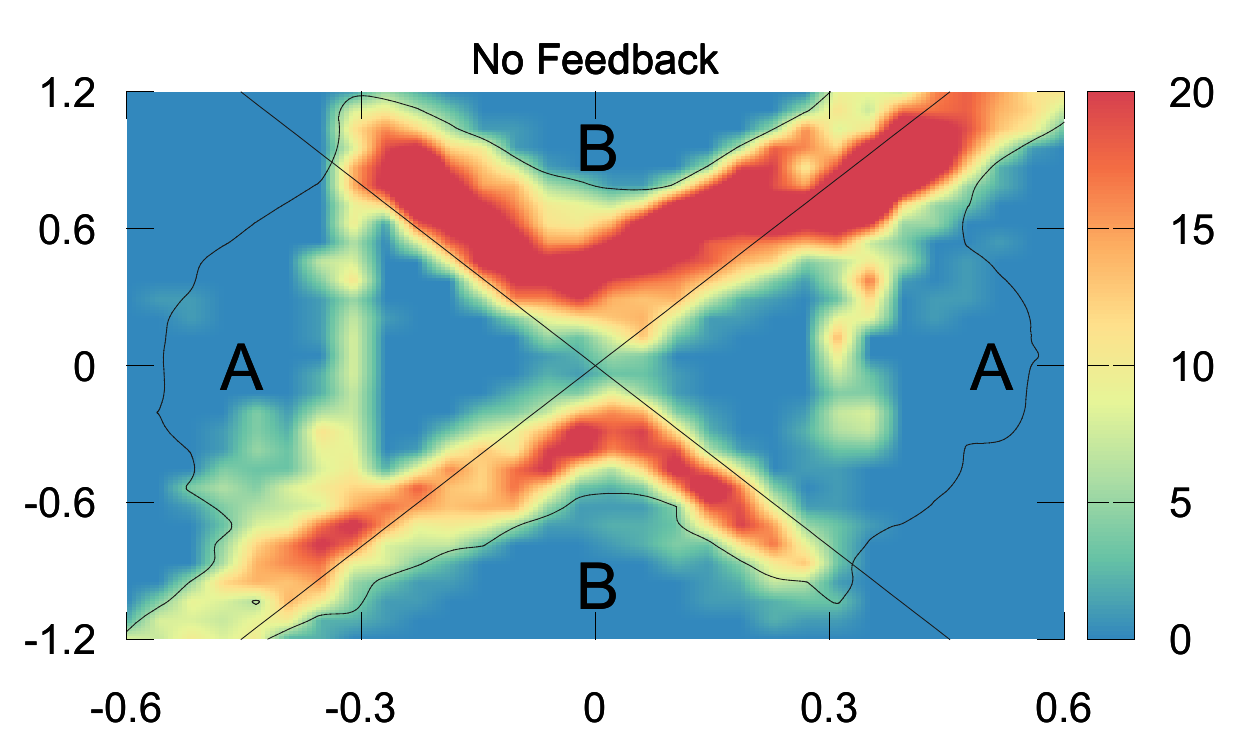}
\includegraphics[width=0.49\textwidth]{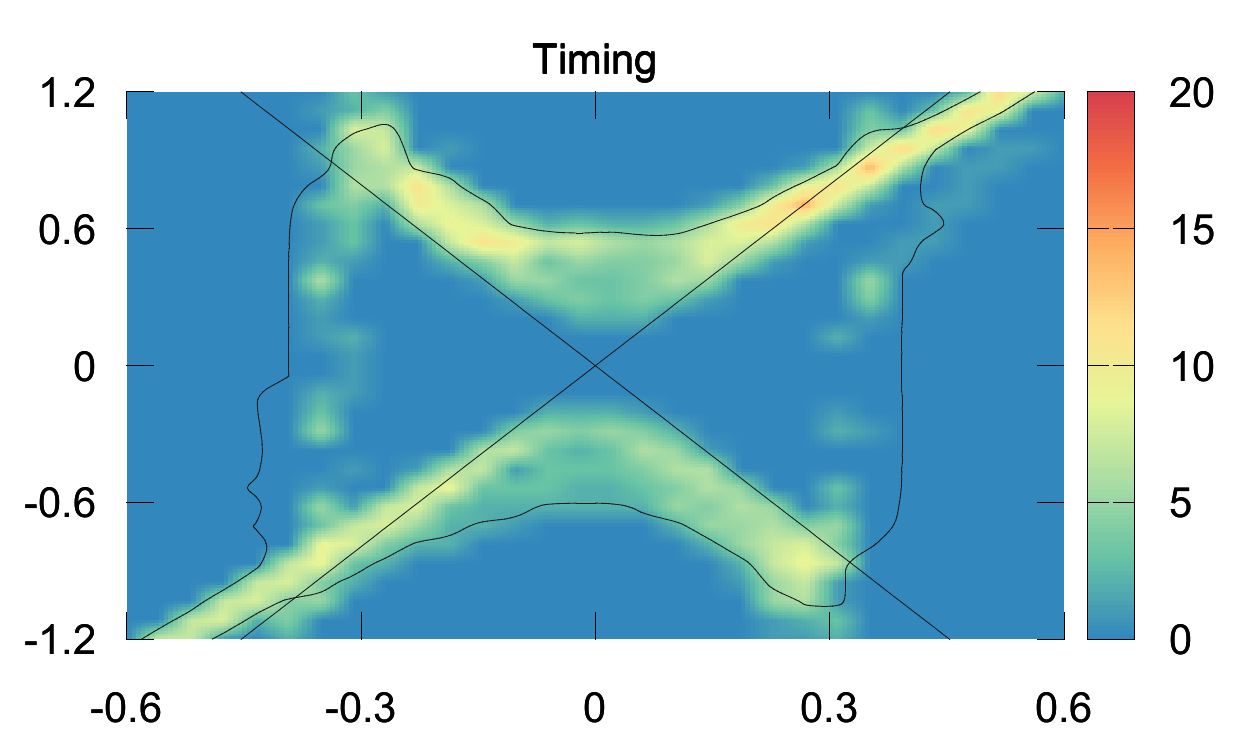}
\includegraphics[width=0.49\textwidth]{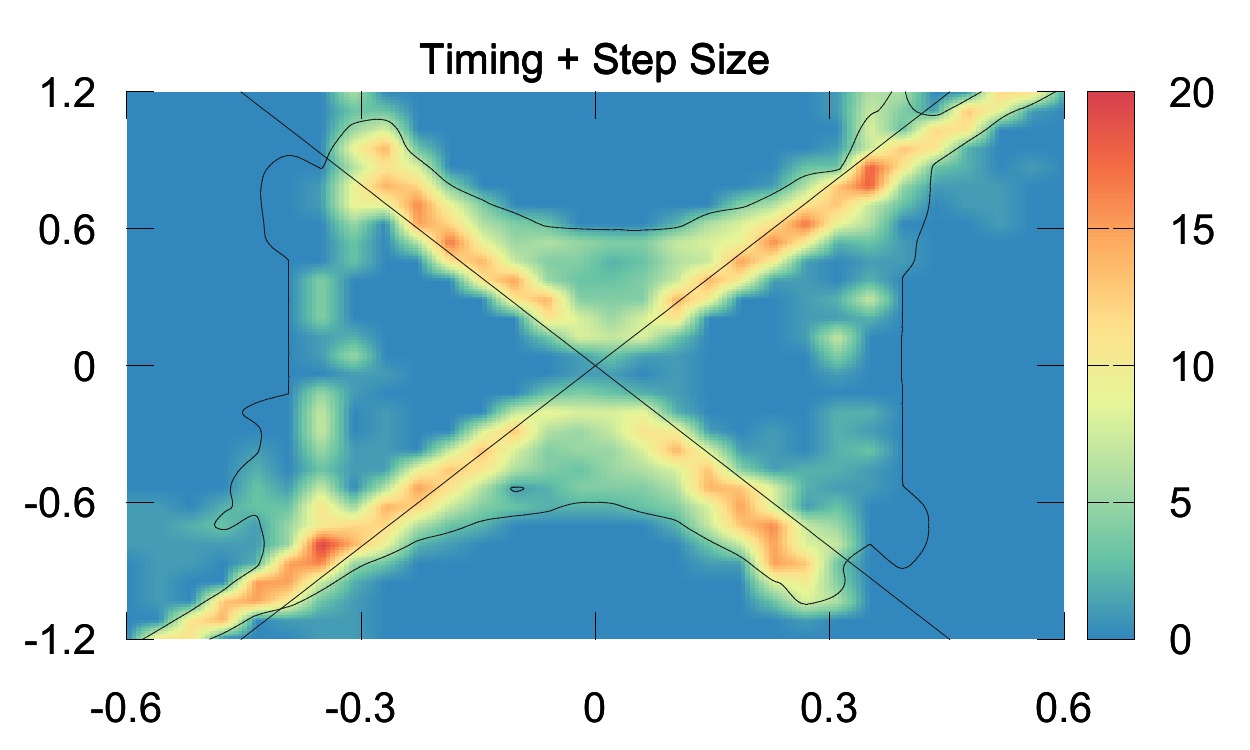}
\includegraphics[width=0.49\textwidth]{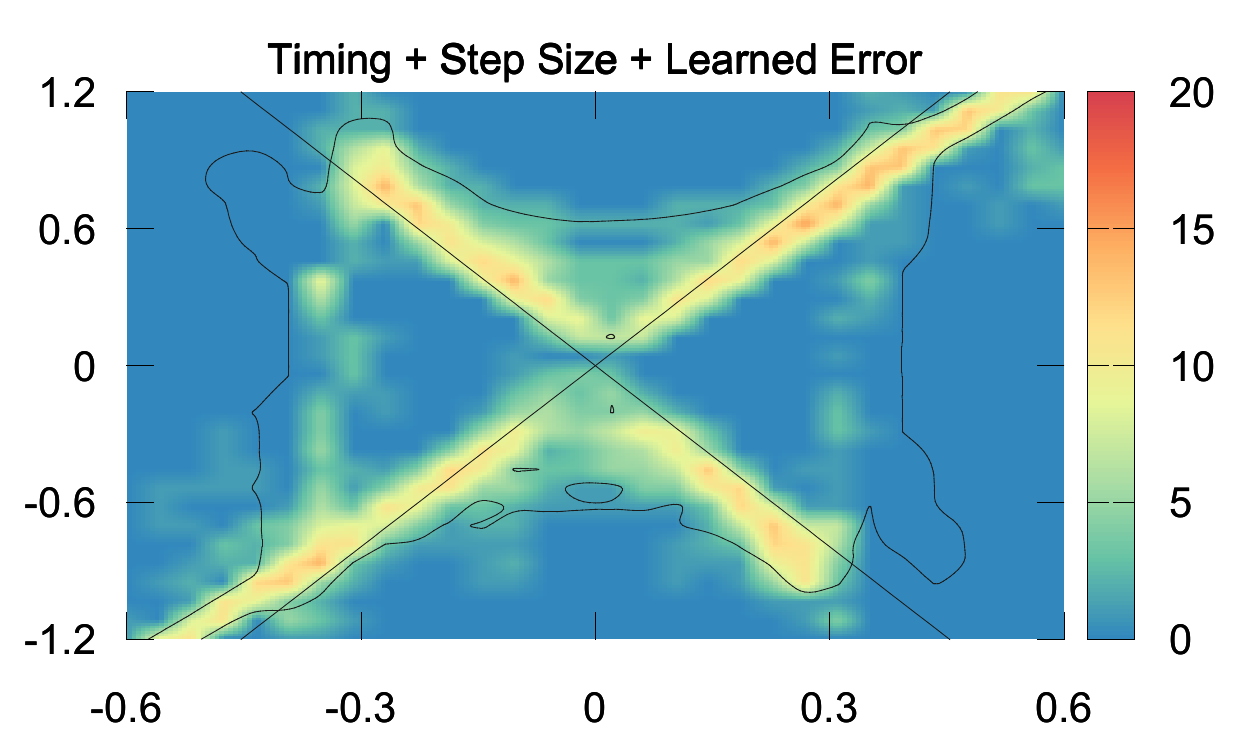}
\vspace*{-2ex}
\caption{Heat maps of unstable regions of the lateral phase space. Color
coding marks the areas that have been crossed by falling trajectories. Thin
black lines contour the cells that were visited at least ten times
during the experiments. Straight zero-energy lines partition the phase space
into stable regions of negative orbital energy (A), and unstable regions of
positive orbital energy (B).}
\label{phasespace}
\end{figure}

For a closer look, Figure~\ref{phasespace} shows heat maps of the lateral phase
space that were generated by backtracking from every fall to the first frame of
a push and incrementing each grid cell that was touched by the center of mass on
the way. The values of the cells are then used for color coding the unstable
regions of the phase space for each controller. The thin black
contours bound the regions of cells that were visited at least ten times during the
experiments. The straight zero-energy lines are computed from the linear
inverted pendulum model that is used to drive the feedback loops. The
zero-energy lines partition the phase space into regions that we would expect to
find based on model assumptions. The areas marked with the letter 'A' are
regions of negative orbital energy. This is where all returning center of mass
trajectories are located and stable lateral oscillations can take place. The
sectors marked with the letter 'B' are of positive orbital energy and contain
state trajectories that will inevitably cross the pivot point and tip over. The
model is reflected by the experimental data, as the vast majority of the states
encountered between a push and a fall are located in the unstable areas of the
heat maps. The fall trajectories of all controllers must originate from the
stable region, since the push is always applied in a stable state of the robot.
The push changes the state trajectory abruptly and transfers it into the
unstable section 'B'. It is evident that the heat map of the open-loop
controller contains a much larger number of falls. The heat maps of the three
feedback controllers look very similar with a strongly reduced number of falls
in comparison with the "No Feedback" experiment. Again, we can conclude that
step timing adaptation plays a pivotal role in preventing a fall.

\begin{figure}[t!] 
\centering
\includegraphics[width=0.9\textwidth]{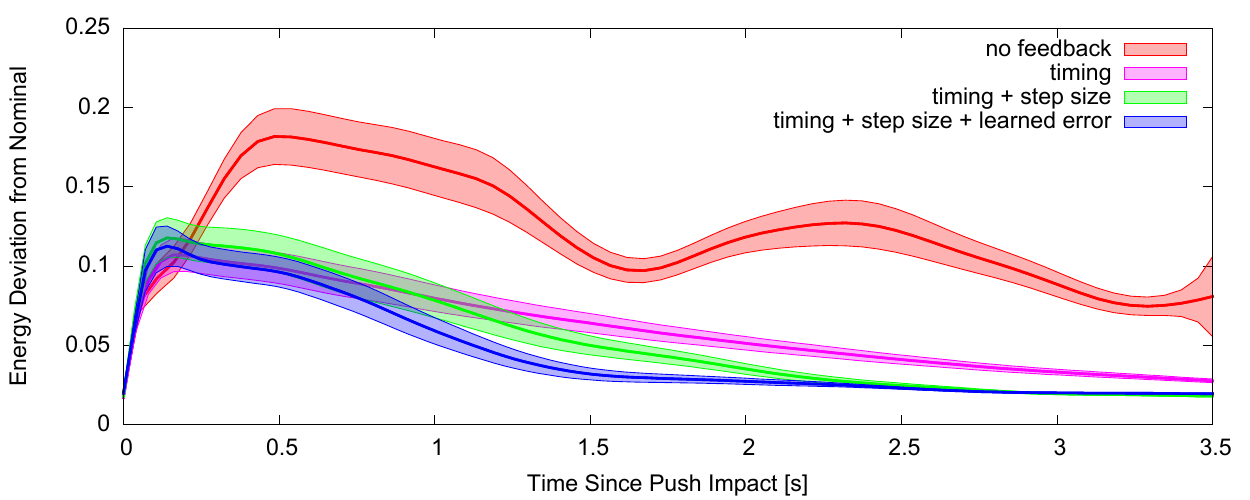}
\includegraphics[width=0.9\textwidth]{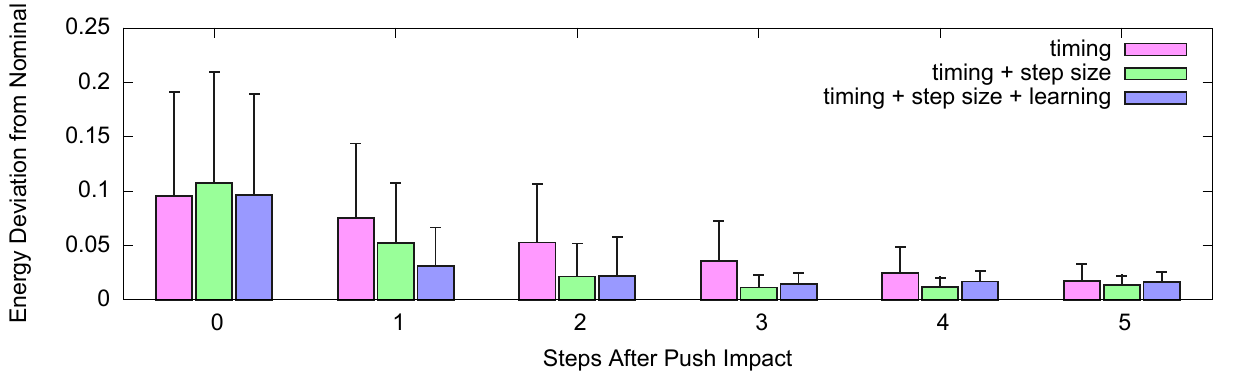} 
\vspace*{-2ex}
\caption{Development of the lateral orbital energy after a push synchronized at
the push impact (top), and at the individual steps after the push (bottom).
While the ``Timing'' controller monotonically returns to a desired level of orbital
energy, the adjustment of step size helps the robot to return to the nominal
energy level much faster. The open-loop controller cannot be sensibly
synchronized with the feedback controllers and thus it has been omitted from the
bottom plot.}
\label{gplot}
\end{figure}

In order to answer the question of how a bipedal walker can benefit from a well
chosen step size, Figure~\ref{gplot} shows the development of the orbital energy
after the disturbance in the cases where the robot did not fall. In the top half
of the plot, the time series of the orbital energy deviation from a nominal
value has been synchronized at the moment of the push impact. Since the
open-loop controller has a tendency to amplify the push impulse, the peak energy
shortly after the push is significantly higher. The wave-like form of the energy
curve suggests that the open-loop controller occasionally disturbs itself. When
using only timing feedback, the disturbance amplification and the self
disturbances disappear and the orbital energy returns monotonically to a desired
level. With the addition of a computed step size, the robot can absorb the
orbital energy much faster. The controller with the learned step size error
shows the best performance in terms of orbital energy dissipation. In the bottom
half of the plot, the energy level with respect to the nominal value has been
synchronized at the individual steps after the push. The fixed-frequency steps
of the open-loop controller cannot be sensibly synchronized with the timed steps
of the feedback controllers and thus have been omitted from the bottom plot. The
first group of boxes show the energy deviation that has been measured during the
step that was pushed. The second group of boxes at the index 1 represent the
``capture step'', the first step after the push. As in theory a full recovery is
possible with one step, the efficiency of the capture step is of particular
interest. The efficiency of a step can be computed as $1 - \frac{e_s}{e_{s-1}}$,
where $e_{s-1}$ and $e_s$ are the excess energy levels before and after the
step. The step efficiency of the step timing controller is 21\%. Adding the step
size modification improves the step efficiency to 51\%, and learning further
increases the energy absorption rate to 68\%. Accelerating the return to a
nominal, stable state has a positive effect on overall bipedal stability. The
walker is ready to face the next disturbance in a shorter amount of time and
thus not only the magnitude, but also the frequency of impulses that the robot
can handle, is increased.

\section{Conclusions}

The TeenSize class experienced an uplift during the 2013 competition. Five teams
were at the competition site and played games with more than one operational
robot on the field from each team. Several technical challenges were completed.
All teams were able to advance their software to cope with the new challenge of
localization with symmetrical landmarks.

In the final, our robots met team CIT-Brains from Japan. In the beginning of the
match, each team played with two players on the field. CIT-Brains played an
offensive strategy with two strikers while team NimbRo designated one player as
goal keeper. The CIT team managed to press onward towards the NimbRo goal, but
the NimbRo robots defended against the attacks reliably. The obstacle avoidance
feature of the CIT robots appeared to be a bit too aggressive and they
approached the NimbRo robots too closely and often stepped on their toes, which
made the CIT robots fall over. NimbRo striker Copedo used the opening gaps to
score. Team NimbRo successfully demonstrated dynamic role assignment that
temporarily assigned the goal keeper Dynaped the striker role when Copedo had to
be taken out of the game. While in the second half, team CIT Brains had to
reduce the number of players to one due to technical difficulties, team NimbRo
managed to maintain two operational players throughout the game and scored
reliably. Consequently, team NimbRo won the finals with a score of 4:0 and
successfully defended its title for the fifth time in a row.

The stability of the gait of our robots and their robustness to disturbances was
one of the key factors for our success. The online learning method outlined in
this work will contribute to even faster stabilization of bipedal walking in
future competitions.

\section{Acknowledgment}

This work is supported by Deutsche Forschungsgemeinschaft (German Research
Foundation, DFG) under grants BE 2556/6 and BE 2556/10.

\bibliographystyle{unsrt}
\bibliography{ms}

\end{document}